# Real-Time Illegal Parking Detection System Based on Deep Learning

*Xuemei Xie [†], Chenye Wang, Shu Chen, Guangming Shi, Zhifu Zhao*

School of Electronic Engineering, Xidian University

xmxie@mail.xidian.edu.cn

## ABSTRACT
The increasing illegal parking has become more and more serious. Nowadays the methods of detecting illegally parked vehicles are based on background segmentation. However, this method is weakly robust and sensitive to environment. Benefitting from deep learning, this paper proposes a novel illegal vehicle parking detection system. Illegal vehicles captured by camera are firstly located and classified by the famous Single Shot MultiBox Detector (SSD) algorithm. To improve the performance, we propose to optimize SSD by adjusting the aspect ratio of default box to accommodate with our dataset better. After that, a tracking and analysis of movement is adopted to judge the illegal vehicles in the region of interest (ROI). Experiments show that the system can achieve a 99% accuracy and real-time (25FPS) detection with strong robustness in complex environments.

## CCS Concepts
• **Information systems**➜**Information systems applications**➜**Decision support systems**➜**Data analytics**.

## 1. INTRODUCTION
Illegal parking may cause a lot of problems such as traffic jam and risks of safety. However, its number has increased considerably in cities in recent years, which impose a heavy burden on its detection. To solve the problem of illegal parking, an accurate and real-time detection system is desired.

Several methods have been proposed to build such a system [1-7], most of which are based on separation of foreground and background. Typically, vehicles are first extracted from background and then are tracked. An alarm will be triggered once a vehicle was found to be stationary for a certain period of time in the ROI. [1]-[3] extracts the foreground objects via sophisticated background modeling strategies. They achieve promising performance in simple traffic environment, but cannot work in crowded scenes. [4] proposes to subtract the background constructed by Gaussian Mixture Model to extract the foreground. And then a vehicle is recognized by detecting its wheels. It effectively separates the foreground and background. However, a vehicle cannot be detected when its wheels are occluded. Wahyono et al. [5] proposes to use background subtraction to obtain candidates of stationary regions, and verify a vehicle by exacting scalable histogram of oriented gradient (SHOG) features followed by a support vector machines (SVM) classification. It copes well with lighting changes, but it is hard to design the SHOG feature, and cannot deal with complex weather conditions. [6] and [7] also falls into this category. However, extracting foreground by background segmentation is easily affected by environments, such as illumination changing and the weather. Besides, occlusion is also difficult to deal with.

In this paper, we propose an accurate and real-time illegal parking detection system based on Single Shot MultiBox Detector (SSD) [8], which can locate and identify objects at the same time without generating region proposal, and has the advantages of high recognition accuracy, high speed and strong robustness. After the detection of vehicles in the input image, all the ones in the ROI will be tracked and counted once it stops moving. To improve the performance, we propose to optimize SSD via adjusting the aspect ratio of default box by K-means to accommodate with the illegal detection problem. Experiments show that we achieve a 99% accuracy and a 25 FPS speed in illegal parking detection. And even with illumination changing and in complex weather, it is still robust and can achieve a high accuracy.

This paper is organized as follows. Section 2 gives a brief introduction of SSD. The flowchart of the proposed system is given in Section 3.1. The improvement of SSD is discussed in Section 3.2. Section 3.3 describes the tracking of vehicles in ROI. The judgment of the motion state of a vehicle is described in Section 3.4. Experimental results are shown in Section 4 and we publish our dataset. Finally, Section 5 concludes our work.

## 2. SINGLE SHOT MULTIBOX DETECTOR
SSD is a method for detecting objects in images using a single deep neural network [9]. It can locate and identify objects in images at the same time without extracting the region proposal, which avoid the disadvantage of background segmentation.

SSD is an end to end training deep neural network. There are a set of default boxes with different aspect ratios at each location in several feature maps which are used to locate and classify the objects in input images. During the training stage, all the default boxes are compared with ground truth. The default boxes which have high IOU with ground truth are treated as positives and the others are treated as negatives. The matched boxes are then trained to obtain their four offsets accounting to its ground truth to match the shapes of objects better. The loss function of SSD is given as follows:

$$L(x,c,l,g) = \frac{1}{N}(L_{conf}(x,c) + \alpha L_{loc}(x,l,g))$$

where N is the number of matched default box for a ground truth. $L_{conf}$ is confidence loss and $L_{loc}$ is localization loss. $x_{ij}^p = \{1,0\}$ is an indicator for matching the $i$-th default box to the $j$-th ground truth box of category $p$. The localization loss is a Smooth L1 loss between the predicted box ($l$) and the ground truth box ($g$) parameters. The parameters of location and category are trained at the same time.

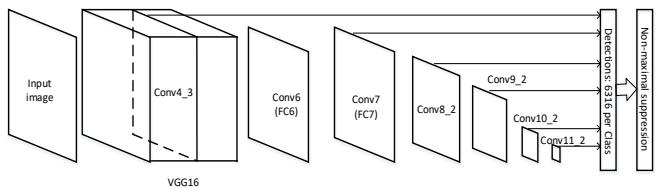

Figure. 1. Single Shot MultiBox Detector network architecture.

The network architecture of SSD is shown in Figure. 1. It is based on a convolutional network such as VGG16 [10], and appends several convolution layers to the end of the base network. All appended feature layer will contribute to the last predictions in the top layer of SSD, so it can naturally predict objects in different scales. In vehicle detection, there are many kinds of vehicles with different scales, and its scale will change with moving. Therefore, it is hard to detect objects with different scales in an image, but SSD can solve the problem well by using feature maps of multiple layers.

SSD is an end to end detector for multiple categories. It can make fast prediction because of the eliminating of bounding box proposals and the subsequent pixel or feature resampling stage. For a 300*300 input image, SSD achieves 74.3% mAP on VOC2007 test at 59 FPS on a NVidia Titian X and 76.9% mAP for 512*512 input, outperforming Faster R-CNN [11] and YOLO [12]. The general framework with high quality and real-time detection results makes it attracting for vehicle detection.

## 3. PROPOSED METHOD
### 3.1 Introduction of The System
As shown in Figure 2, our system mainly consists of four parts: video capturing, SSD-based vehicle detection, tracking and the analysis of movement. The input of the system can be a real-time signal come from a surveillance camera or a local video. Since the illegal regions in different scenarios varies, the ROI need to be marked manually.

Flowchart of the proposed system is shown in Figure 3. For a given frame in a video, firstly we input the image into a trained SSD network to detect the location and category of all vehicles contained in the image by setting a suitable hyper-parameter. And then the vehicles will be tracked and its motion state will be analyzed if it is in the ROI. If it is found that a vehicle is in a quiescent state, the system will start to time it. And if the time before the state changes exceeds a given threshold, the vehicle will be identified as illegal parking and the alarm will be triggered. In order to detect new arrivals within the ROI, and prevent tracking drift problem, the system re-detects vehicles in the video every 25 frames. The newly detected box will be matched with the original box by calculating intersection-over-union (IOU) and inherits the timing information of the old box if matched.

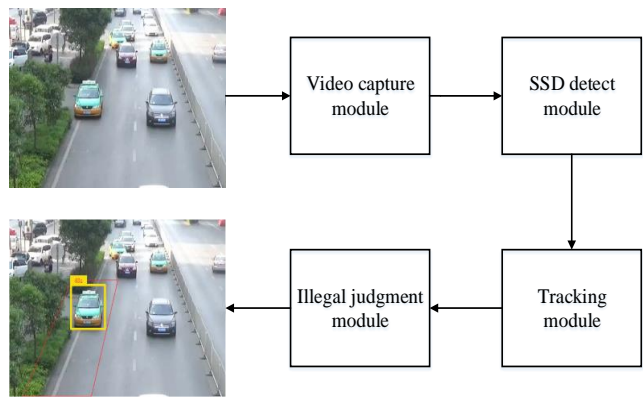

Figure 2. The composition of the proposed system.

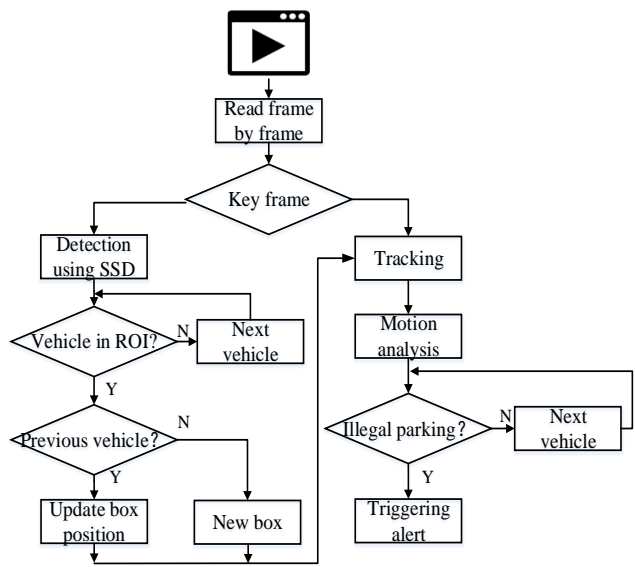

Figure 3. Flowchart of the proposed system.

### 3.2 Improvement of SSD
The original SSD is a network architecture designed for VOC dataset which have objects with different scales and shapes. In other words, it is a general detector. For an illegal parking system, however, the targets are only vehicles which have more single scales and aspect ratios. Big boxes and flat boxes won't match any ground truth box, they are useless for us and increase the computational complexity. An unsuitable default box may bring a lot of background noise. All of these default boxes will reduce the accuracy. It is clear that if we train the original SSD network with our own dataset directly, both speed and accuracy of network will be unsatisfactory. So, the network must be improved to accommodate with our dataset better.

In order to make SSD more suitable for our data set, we redesign the aspect ratios of the default boxes in original SSD. We propose to use k-means to improve SSD. K-means is a method to cluster the input data according to a characteristic of it. We use aspect ratios of all ground truth in dataset as input data. By setting a value of clustering centers K, We have got K clustering centers which are most closely matched with our data sets. One question needs to be notice that these clustering centers cannot be too close, otherwise every default boxes will match the same ground truth. The roles of these boxes is completely repeated and the amount of calculations has increased. The value of K is set according to the distribution of the aspect ratios of the ground truth in the data set. Then we set the aspect ratios of default boxes to obtained ones. And after that we fine-tune the network based on original network parameters by our own dataset.

In our implementation, K is set to 2. Accounting to the result of K-means, we set the aspect ratios to 0.5 and 0.7, respectively. The aspect ratios of original SSD are 0.3, 0.5, 1, 2, and 3, respectively. After trained by our dataset, the optimized model for vehicle detection is obtained. It has better performance for vehicle detection because the aspect radios of default boxes match our dataset better. The accuracy increased from 75% to 77%. The detect results are shown in Figure 4.

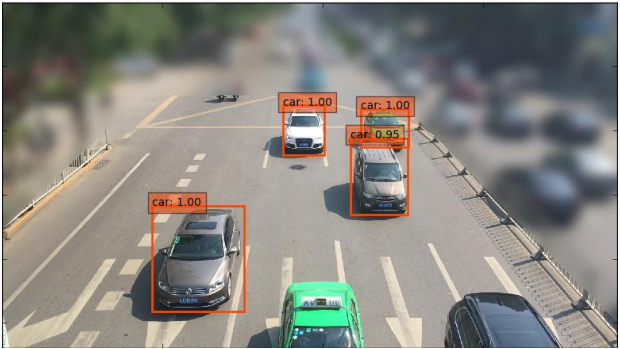

(a)  Detection in sunny weather

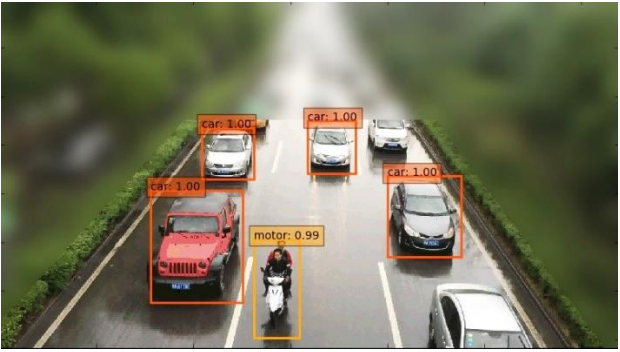

(b)  Detection in rainy weather

Figure 4. The detection of optimized SSD

### 3.3  Tracking

The vehicles in ROI will be tracked after detection with the improved SSD model. Here we use the template matching to track the vehicles. In order to speed up, templates are converted to gray scale. At first, the detection results are selected as the original template, which is denoted as $T(x, y)$. After the arrival of a new frame, the entire image is searched to find out the best matched area with the template. For each candidate area $I(x, y)$, we evaluate its score $R(x, y)$ with $T(x, y)$, as defined in Eq. (1). The region with the highest score is identified as the new location of the object.

$$R(x,y) = \sum_{x',y'}\bigl(T'(x',y') \times I'(x+x',y+y')\bigr). \quad (1)$$

$T'(x, y)$ and $I'(x, y)$ are the normalized image block of $T(x, y)$ and $I(x, y)$, respectively. The normalization is established by subtracting the respective mean values and dividing them by the respective variance, which are given by

$$T'(x,y) = \frac{T(x,y) - \frac{1}{w \times h}\sum_{x',y'} T(x',y')}{\sqrt{\sum_{x',y'} T(x',y')^2}}$$

and

$$I'(x,y) = \frac{I(x,y) - \frac{1}{w \times h}\sum_{x',y'} I(x',y')}{\sqrt{\sum_{x',y'} I(x',y')^2}}$$

After that, the normalized images can reduce the affection of illumination changing. $R(x, y)$ evaluates the degree of similarity between the original template and the target one. The final result is normalized to -1 to 1, in which 1, -1 and 0 means that the two images are perfectly matched, exactly the opposite, and no linear relationship between the two templates, respectively.

### 3.4  Analysis of Illegal Parking

In order to determine whether a vehicle has an illegal behavior, the system analyzes the state of the vehicles being tracked during the whole tracking process. The distance that the vehicle moves in the region of interest is obtained by making the difference between the positions of the adjacent two frames. If the difference between the two positions is greater than a given threshold, it is considered to be moving. Otherwise, it is in a quiescent state, and the stop time of the vehicle is counted at the same time. The alarm will be triggered when the stop time of any vehicle exceeds the given threshold.

One problem needs to be noticed that a new vehicle may enter into ROI at any moment. So the system detect every certain frames. It can prevent tracking drift problem at the same time. Since the SSD will always detect all the vehicles in an input image including the vehicles is being tracked, so we need to match the newly detected boxes with the boxes in last frame. Here we match two boxes by calculating the IOU between them. If an IOU value of two boxes is greater than the given threshold, we think the two boxes contain the same vehicle and the new box will inherit the timing information of the old box.

## 4. EXPERIMENTAL RESULTS

### 4.1  Implementation Setting

SSD model is trained by our own dataset [13] which contains various kinds of traffic conditions including sunny, raining, shadow, crowd and smoothness. Our illegal parking detection system is implemented by C/C++ programming language. In order to synchronize the detection and tracking module, we use two threads for them. The system runs on the PC equipped with Core i7-4790 CPU, at 3.60GHz frequency and 8GB RAM. In our experiments, the confidence threshold of SSD is set to 0.6. Once a vehicle stops for 15s, it will be identified as illegal parking, and system will begin to time for illegal vehicles until they moving.

### 4.2  Detection Results

Our system is evaluated on our own dataset described above. Our system can detect cars trucks and motorcycles. In this paper we only consider cars. The test results are shown in Figure 4. The red box represents the ROI which is drawn according to forbidden area. The yellow boxes represent illegal parking vehicles detected by the system. The number on the yellow box records the illegal parking time.

The detection process of the system in sunny days is presented by frame 330 to frame 1238 in Figure 5. There are three cars in the ROI in frame 330. As the two black cars enter into the ROI earlier than the white one, the two black cars are identified as illegal vehicles earlier in frame 467. One of the black car start to move in frame 697, so the system removes its illegal state. The stop time of the white car increases to 15s in frame 875, and thus it is identified as an illegal vehicle. The white car starts to move in the following frames. Another black car is the only illegal car kept to the end. Figure 6. shows the detection process in rainy day. This system achieved a 99% detection accuracy on our own test videos.

## 5. CONCLUSION

This paper proposes a deep learning based framework to detect illegal parking in ROI. It achieves high accuracy and real-time detection results. Besides, our system has a high degree of robustness and can adapt to a variety of complex environments.

## 6. ACKNOWLEDGMENTS

This work is supported by Natural Science Foundation (NSF) of China (61472301).

## 7. REFERENCES

[1] E. Herrero-Jaraba, Orrite-Uru, C. Uela, and J. Senar, "Detected motion classification with a double-background and a neighborhood-based difference," Pattern Recognition Letters, vol. 24, pp. 2079-2092, 2003.

[2] L. Maddalena and A. Petrosino, "Stopped Object Detection by Learning Foreground Model in Videos," IEEE Transactions on Neural Networks & Learning Systems, vol. 24, pp. 723-735, 2013.

[3] P. Fatih and H. Tetsuji, "Robust Abandoned Object Detection Using Dual Foregrounds," EURASIP Journal on Advances in Signal Processing, vol. 2008, pp. 1-11, 2007.

[4] C. Mu, M. Xing, and P. Zhang, "Smart Detection of Vehicle in Illegal Parking Area by Fusing of Multi-features," in *International Conference on Next Generation Mobile Applications, Services and Technologies*, pp. 388-392, 2015.

[5] Wahyono, A. Filonenko, and K. H. Jo, "Illegally parked vehicle detection using adaptive dual background model," in *Industrial Electronics Society*, pp. 25-28, 2015.

[6] S. Banerjee, P. Choudekar, and M. K. Muju, "Real time car parking system using image processing," in *International Conference on Electronics Computer Technology*, pp. 99-103, 2011.

[7] W. Wahyono and K.-H. Jo, "Cumulative Dual Foreground Differences For Illegally Parked Vehicles Detection," *IEEE Transactions on Industrial Informatics,* pp. 1-9, 2017.

[8] W. Liu, D. Anguelov, D. Erhan, C. Szegedy, S. Reed, C.-Y. Fu*, et al.*, "SSD: Single shot multibox detector," in *European Conference on Computer Vision*, pp. 21-37, 2016.

[9] C. Mu, M. Xing, and P. Zhang, "Smart Detection of Vehicle in Illegal Parking Area by Fusing of Multi-features," in *International Conference on Next Generation Mobile Applications, Services and Technologies*, pp. 388-392, 2015.

[10] K. Simonyan and A. Zisserman, "Very deep convolutional networks for large-scale image recognition," *arXiv preprint arXiv:1409.1556,* 2014.

[11] R. Girshick, "Fast R-CNN," *Computer Science* pp. 1440-1448, 2015.

[12] J. Redmon, S. Divvala, R. Girshick, and A. Farhadi, "You only look once: Unified, real-time object detection," in *Proceedings of the IEEE Conference on Computer Vision and Pattern Recognition*, pp. 779-788, 2016.

[13] http://see.xidian.edu.cn/faculty/gmshi/dataset.htm

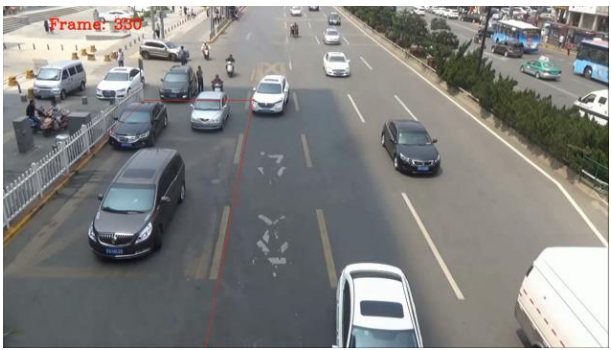

Frame 330

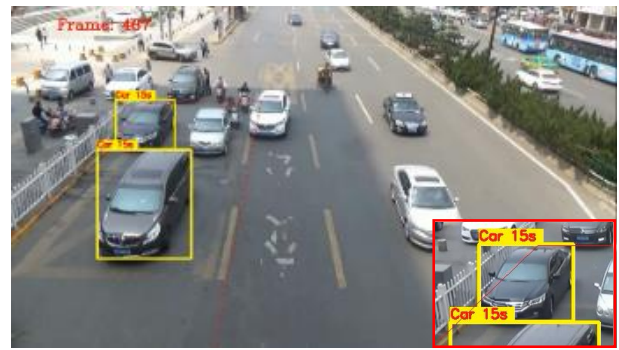

Frame 467

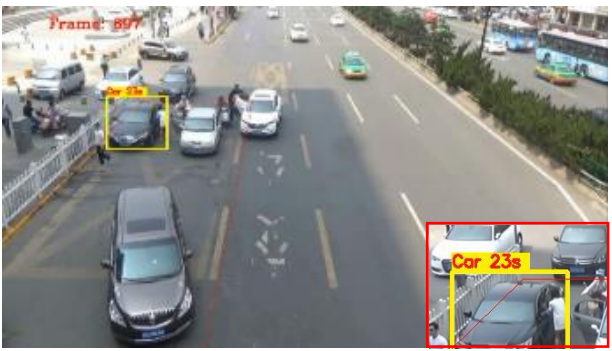

Frame 697

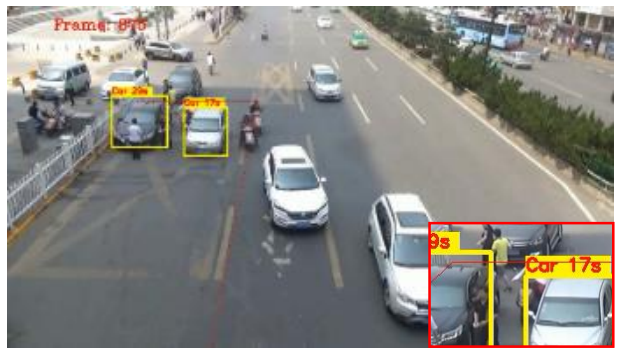

Frame 875

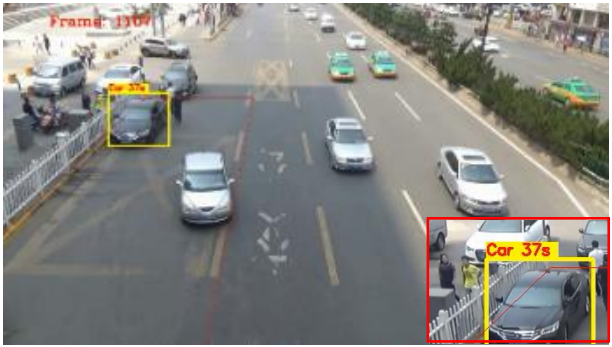 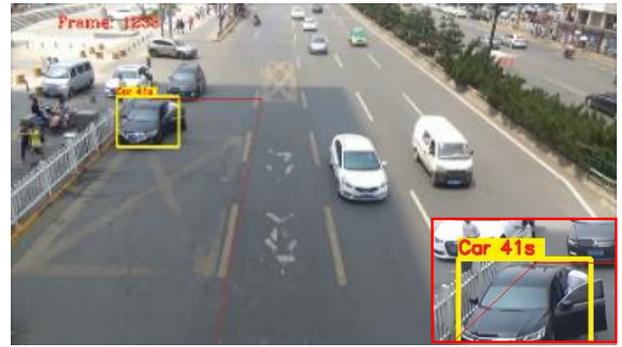

Frame 1107 | Frame 1238

Figure 5. Detection in the case of sunny weather. There are three vehicles in ROI in frame 330. The two black cars are firstly identified as illegal vehicles according to their parking time in frame 467 and the white car follows in frame 875. When one of black car and the white car start to move, the system removes their illegal state. The black car is the only illegal car in the end in frame 1238.

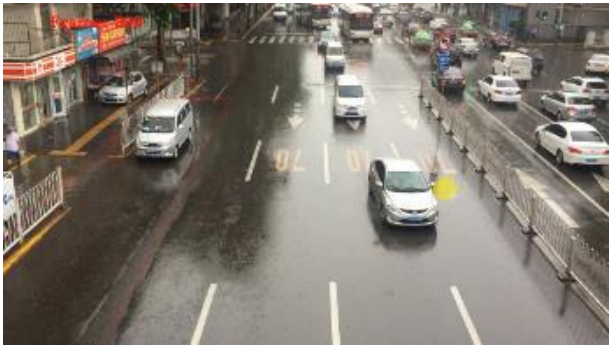 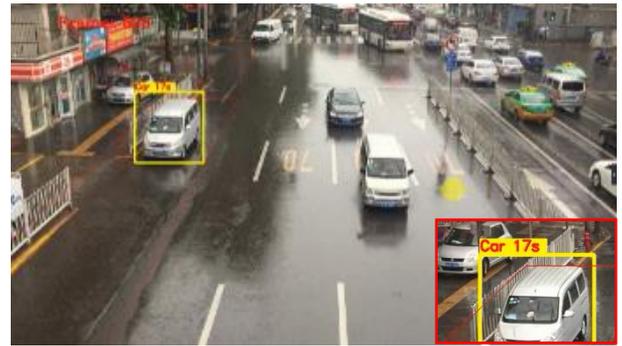

Frame 399 | Frame 528

Figure 6. Detection in the case of rainy weather

5